# Convolutional Neural Networks in Multi-Class Classification of Medical Data

YuanZheng Hu
EECS, University of Ottawa

*bhu078@uottawa.ca*

Marina Sokolova
IBDA@Dalhousie University and University of Ottawa

*sokolova@uottawa.ca*

**Abstract** We report applications of Convolutional Neural Networks (CNN) to multi-classification classification of a large medical data set (*Diabetes 130-US hospitals for years 1999-2008 dataset*). We work on multi-classification of the patient's readmission, which resulted in classification of three classes: *0*, *<30*, or *> 30* days. We discuss in detail how changes in the CNN model and the data pre-processing impact the classification results. In the end, we introduce an ensemble model that consists of both deep learning (CNN) and shallow learning models (Gradient Boosting). The method achieves Accuracy of 64.93%, the highest three-class classification accuracy we achieved in this study. Our results also show that CNN and the ensemble consistently obtain a higher Recall than Precision. The highest Recall is 68.87, whereas the highest Precision is 65.04.

## 1. Introduction

Deep Learning algorithms are the state-of-art AI techniques that use self-optimization through learning (LeCun et.al, 2015; O'Shea & Nash, 2015). Deep Learning algorithms became major investigative tools in life and social science, with health science and medical science significantly benefitted from their applications (Holzinger et al, 2017). Deep Learning methods are based on neural networks, a representation of human neurons in the form of abstract data structure. Whereas a basic classical neural network consists of three layers (the input layer, the hidden layer, and the output layer), the term "deep" often refers to multiple hidden layers. The learning results depend on many factors, including the network architecture, optimization algorithms, the selected loss function. Data preprocessing also plays an important role in the obtained results. This includes feature selection and construction, instance sample selection, including over- and under-sampling of imbalanced data.

In this work, we report applications of Convolutional Neural Networks (CNN) to multi-classification classification of a large medical data set[1]. The current work continues our previous work on explainable multi-class classification of medical data (Hu and Sokolova, 2020). As in the previous work, we gather empirical evidence on *Diabetes 130-US hospitals for years 1999-2008 dataset*. The set and its features were introduced by Strack et.al. (2014) and deposited to the UCI Machine Learning Repository. The dataset has 100000 instances and 55 features. [2] The data processing, including noise removal and knowledge-based feature construction, has been presented and discussed in detail in (Hu & Sokolova, 2020).

We work on multi-classification of the patient's readmission, which resulted in classification of three classes: *0*, *<30*, or *> 30* days. This is a novel task that provides more insights into patient readmission

---

[1] Our results with Recurrent Neural Networks (RNN) are reported in Appendix 2.
[2] UCI Machine Learning Repository: Diabetes 130-US hospitals for years 1999-2008 Data Set

prospects than the binary classification of patients' readmission. The results of the three-class classification can lead to finer-grained patient-centric and privacy protection analysis of the data.

We discuss in detail how changes in the CNN model and the data pre-processing impact the classification results. To select the best model, we compute Macro F-score, Precision, Recall, and Accuracy (Sokolova & Lapalme, 2009). In the end, we introduce an ensemble model that consists of both deep learning (CNN) and shallow learning models (Gradient Boosting). The method achieves Accuracy of 64.93%, the highest three-class classification accuracy we achieved in this study. Our results also show that CNN and the ensemble consistently obtain a higher Recall than Precision. The highest Recall is 68.87, whereas the highest Precision is 65.04.

In this work, we use software supported by Scikit-Learn and PyTorch. Data analysis and processing libraries include Pandas, Matplotlib, and NumPy. All coding and model training is done by using CPU I7-9700K and GTX 1060.

## 2. Convolutional Neural Networks

### 2.1 Preliminaries

Convolutional Neural Networks (CNN) consist of convolutional layers, pooling layers and fully-connected layers (O'Shea &Nash, 2015). The key factors of CNN applications can be listed as follows: local connections, shared weights, pooling, and the use of the layers. The idea behind location connections is based on certain data sets, e.g., images and signal sequences, being highly correlated within the same location; thus, the feature maps may exhibit the same motif. For each round of CNN application, the input data are broken into several feature maps. Each of the feature maps can be treated as the input of a layer, with a convolutional layer followed by a pooling layer taking in the inputs to process the features. Different from regular neural networks - where each neuron is associated with one weight and connected to the next - each convolutional layer shares an array of weights called *filter*. The weighted sum from the filter and the convolutional layer is then passed into a non-linearity function and then fed into the pooling layer.

The pooling layer, used after the convolutional layer, merges the feature maps with similar features. (The feature maps that are close to each other are semantically identical, therefore the original inputs can be condensed into a smaller set of features.) A common CNN architecture has a few stacked convolutional layers and pooling layers. Their outputs are forwarded to the fully-connected layers. In the end, the weights in filters will be trained by the backpropagation. In practice, there will be another parameter called *out_channels* which determines the number of outputs at each round of convolution; the parameter helps the networks to generate multiple outputs based on different choices of weights instead of training based on one weight.

### 2.2 Vanilla CNN

We start with a vanilla CNN network, that incorporates CNNs as its layers. The network consists of 6 layers, with 2 layers of CNN, 1 layer of max pooling, and 3 forward layers. Architecture is shown in Figure 1; our configuration parameters are shown in Table 1.

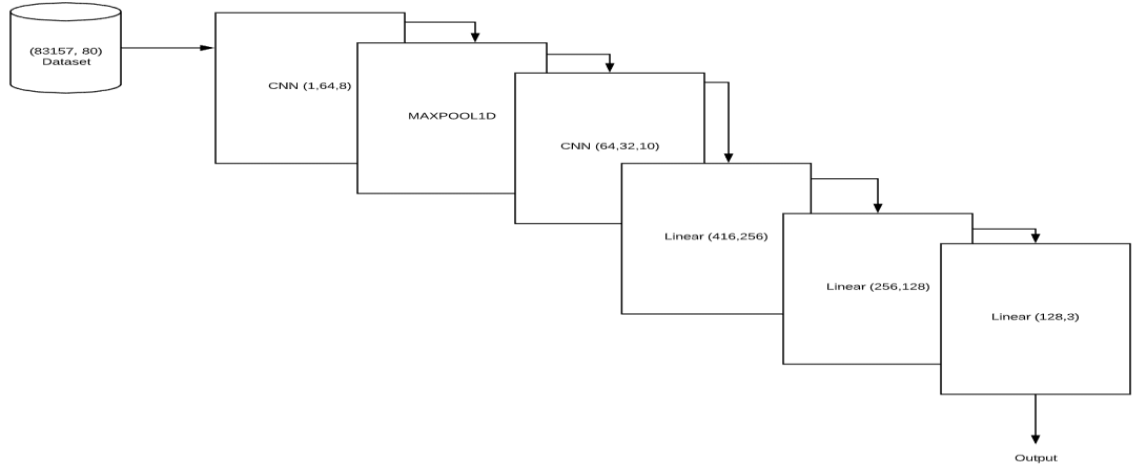

*Figure 1, Architecture of the vanilla CNN*

We use cross-entropy loss function as our criterion to decide on the predicted label. The cross-entropy loss function in Pytorch combines the negative log likelihood loss and log softmax function, and outputs a numeric probability value on different classes. We use the largest value as our predicted label. Note that the negative log function is more sensitive when the confidence of a classified class is low. In the other words, the negative log function will give more penalties when the confidence of classification is low; this can be one of the advantages of using the Cross-Entropy loss function that consists of negative log-likelihood and log softmax function. The formula of Cross-Entropy Loss function used in this report is illustrated below

*Equation 1. Cross-Entropy loss function*

$$loss(x, class) = -\log \left(\frac{\exp(x|class)}{\sum_j \exp(x_j)}\right)$$

We opted for Adaptive Moment Estimation (Adam) (Kingma et.al., 2014) as our initial optimizer, since it performs well when dealing with large datasets. The Adam algorithm is outlined as follows: for each parameter, the optimizer will first take the exponential average of the square of gradients and use that value along with the learning rate to define the step size on the weights of the neural networks. In the last step, it adds the modification of the weights to the original weights.

To make sure our training set is unbiased towards the class distribution, we use 10-fold cross-validation with a stratified shuffle. The vanilla CNN has a learning rate of 0.0001, trained on 64 batches and 10 epochs. To further adjust our vanilla model, we will perform parameter tuning on it. Since we found the learning rate and batches have an impact on accuracy, we will consider epochs, learning rate, and batches as parameters to tune. For each configuration, we will perform 10-fold cross-validation and take the average accuracy for each configuration to compare. Total training time for this step took around 16 hours to finish. We will test each combination of configuration, with epochs value 10,50, learning rate 1e-05, 1e-04, 0.001, 0.01, and batch size 16,32,64. Table 1 reports on Accuracy, Macro Recall, Precision, and F-score. The highest obtained Accuracy is 62.82%.

*Table 1: Results of the vanilla CNN runs; the best results and the configuration are in **bold**.*

| Epochs | Learning rate | Batch size | Accuracy | Recall | Precision | F1 |
|---|---|---|---|---|---|---|
| 10 | 1e-05 | 16 | 60.71 | 65.00 | 60.71 | 62.77 |
| 10 | 1e-04 | 16 | 61.48 | 67.50 | 61.48 | 64.35 |
| 10 | 0.001 | 16 | 53.75 | 56.31 | 53.75 | 55.00 |
| 10 | 0.01 | 16 | 33.33 | 33.33 | 33.33 | 33.33 |
| 10 | 1e-05 | 32 | 60.14 | 63.08 | 60.14 | 61.57 |
| 10 | 1e-04 | 32 | 61.68 | 62.98 | 60.57 | 61.75 |
| 10 | 0.001 | 32 | 59.52 | 64.97 | 59.64 | 62.19 |
| 10 | 0.01 | 32 | 33.33 | 33.33 | 33.33 | 33.33 |
| 10 | 1e-05 | 64 | 59.49 | 62.37 | 59.49 | 60.90 |
| 10 | 1e-04 | 64 | 60.57 | 62.98 | 60.57 | 61.75 |
| 10 | 0.001 | 64 | 59.64 | 64.97 | 59.64 | 62.19 |
| 10 | 0.01 | 64 | 33.33 | 33.34 | 33.33 | 33.33 |
| 50 | 1e-05 | 16 | 62.69 | 66.54 | 62.69 | 64.56 |
| 50 | 1e-04 | 16 | 61.02 | 64.04 | 61.02 | 62.49 |
| 50 | 0.001 | 16 | 49.61 | 50.86 | 49.60 | 50.22 |
| 50 | 1e-05 | 32 | 62.62 | 67.06 | 62.62 | 64.77 |
| 50 | 1e-04 | 32 | 62.77 | 67.07 | 62.77 | 64.85 |
| 50 | 0.001 | 32 | 61.54 | 66.65 | 61.54 | 63.99 |
| 50 | 0.01 | 32 | 33.33 | 33.33 | 33.33 | 33.33 |
| 50 | 1e-05 | 64 | 62.26 | 66.66 | 62.26 | 64.39 |
| **50** | **1e-04** | **64** | **62.82** | **67.69** | **62.81** | **65.16** |
| 50 | 0.001 | 64 | 62.02 | 67.21 | 62.02 | 64.51 |
| 50 | 0.01 | 64 | 33.32 | 33.32 | 33.32 | 33.32 |

We make a few adjustments to our vanilla CNN. We hypothesize that a deeper network may result in better accuracy, thus we add CNN to the vanilla architecture (Figure 2). Then we set the window size to 4, to give more information to the multinetwork player in the front. At last, we will add dropout layers to each linear layer in the front; dropout layer will randomly stop update the weights for a few random weights, we set the threshold to be 0.2 in this case. The best obtained Accuracy is 62.91%.

*Figure 2,* Adjustments of the vanilla CNN architecture (CNN 2.0)

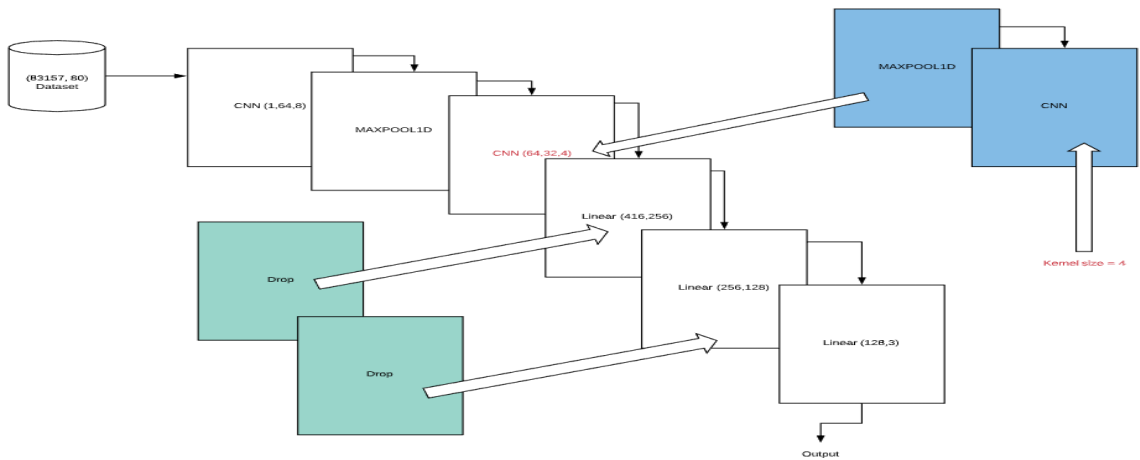

## 2.3 Feature Selection

We investigate the impact of univariate feature selection on the classification performance.

We first perform Chi-square statistical test on the features and the target label. For each feature we will using a Chi-square test to find if there is dependency with the target, and then rank the features according to their score. We use the SelectionKBest method from Scikit-Learn. We select the top 16,32, 58 features for CNN, Gradient Boosting, and Random Forest. The latter two algorithms have achieved the best classification in (Hu & Sokolova, 2020). Features are normalized between 0 to 1 to perform this test since the negative value is not permitted here. However, the Chi-square feature selection does not improve on the accuracy of classification. We drop 16 features from further tests as they provided for the lower accuracy.

We then use Pearson correlation to compute the correlation between each feature and the target, then rank their score and pick the top 20 and 58 features to train the dataset. Features with zero mean value are excluded from this test. In this case, Gradient Boosting has an insignificant increase from 64.75% to 64.83%.

For the ANOVA test (F-value test), we propose null hypnosis for each feature and the target label, then find F-value, rank the features, and train the dataset. We pick the top 20 and 58 features to perform the test. Features with zero variance are excluded from this test. In this case, there is an observable decrease for CNN when 20 features are used. All the results are summarized in Table 2.

*Table 2, Accuracy results after univariate feature selection.*

|  | CNN | Gradient Boosting | Random Forest |
|---|---|---|---|
| All features | 62.91% | 64.75% | 64.68% |
| Chi-Square test | | | |
| 16 features | 57.20% | 62.93% | 61.51% |
| 32 features | 61.12% | 64.09% | 63.58% |
| 58 features | 62.63% | 64.47% | 64.43% |
| Pearson correlation test | | | |
| 20 features | 58.92% | 64.12% | 63.19% |
| 58 features | 62.20% | 64.83% | 64.59% |
| ANOVA test | | | |
| 20 features | 58.40% | 64.10% | 63.50% |
| 58 features | 62.69% | 64.92% | 64.72% |

We report in detail CNN results after the feature selection. All tests are done with 10-fold stratified cross validation on CNN. We also investigate the impact of the 23 medication features (Appendix 1) on the CNN performance; the features had been shown to improve Macro Recall of the patient readmission classification (Hu & Sokolova, 2020).

*Table 3: CNN Accuracy (%) after feature selection: with and without 23 medication features*

| Feature selection | With 23 features | Without 23 features |
|---|---|---|
| Anova top 16 features | 57.69 | 55.56 |
| Anove top 20 features | 58.40 | 58.57 |
| Anova top 40 features | 61.31 | 62.51 |
| Anova top 58 features | 62.69 | 62.92 |
| Correlation top 20 features | 58.92 | 59.08 |
| Correlation top 40 features | 61.89 | 62.33 |
| **Correlation top 58 features** | **62.20** | **63.17** |
| Chi2 top 16 features | 57.20 | 58.81 |
| Chi2 top 32 features | 61.12 | 61.72 |
| Chi2 top 58 features | 62.64 | 63.06 |

## 2.4. Oversampling and under sampling methods

To balance the original Diabetes data set, we have been using SMOTE are our oversampling method (Hu & Sokolova, 2020). We performed a few other oversampling methods: ADASYN, Random oversampling, SVMSMOTE, and BorderSMOTE, and an undersampling NearMiss. We adjusted our dataset using these four oversampling methods and found that ADASYN performs the best with an accuracy of 64.40%.

| Method | With 23 features | Without 23 features |
|---|---|---|
| ADASYN | 63.70 | **64.41** |
| Random oversampling | 49.85 | 49.39 |
| SVMSMOTE | 63.03 | 63.25 |
| BorderSMOTE | 62.64 | 63.17 |
| NearMiss | 57.33 | 57.41 |

*Table 4, Accuracy (%): comparison of four oversampling methods and an under-sampling method*

As shown in the table above, ADASYN provided for the best accuracy among all the oversampling and an under-sampling method. Thus, the following sections will be based on the oversampling method of ADASYN. Our benchmark accuracy is established as 64.41%.

## 2.5 CNN 2.0 adjustment

We work with CNN 2.0 presented in Fig. 2.

*Step 1:*

We explore whether a more complex network might boost up the accuracy. In the new architecture, the last three layers contain neurons 2560, 1280, and 3, respectively. The ad-hoc layer parameters do not improve on the CNN 2.0 performance. The results are given in Table 5.

| Model | Accuracy | Recall | Precision | F1 |
|---|---|---|---|---|
| Layers with more neurons | 60.83% | 61.71% | 60.46% | 61.08% |

*Table 5, Adjusted CNN 2.0 accuracy, step 1*

*Step 2:*

Next, we are going to explore the impact of the kernel size on the accuracy; from the current 4, we decrease it to 1 and 2 and increase it to 6. As we can observe from the tests (Table 6), filter size 4 already gives us the best result.

| Model | Accuracy | Recall | Precision | F1 |
|---|---|---|---|---|
| Filter size 1 | 63.63 | 67.75 | 63.47 | 65.54 |
| Filter size 2 | 64.25 | 67.98 | 64.10 | 65.98 |
| Filter size 4 (Current) | **64.41** | **68.28** | **64.26** | **66.21** |
| Filter size 6 | 63.85 | 67.80 | 63.70 | 65.69 |

*Table 6, Adjusted CNN 2.0 accuracy (%), step 2*

*Step 3*

To further explore the impact of kernel size, we made a deeper and more complex networks, as shown by Fig 3. [3]

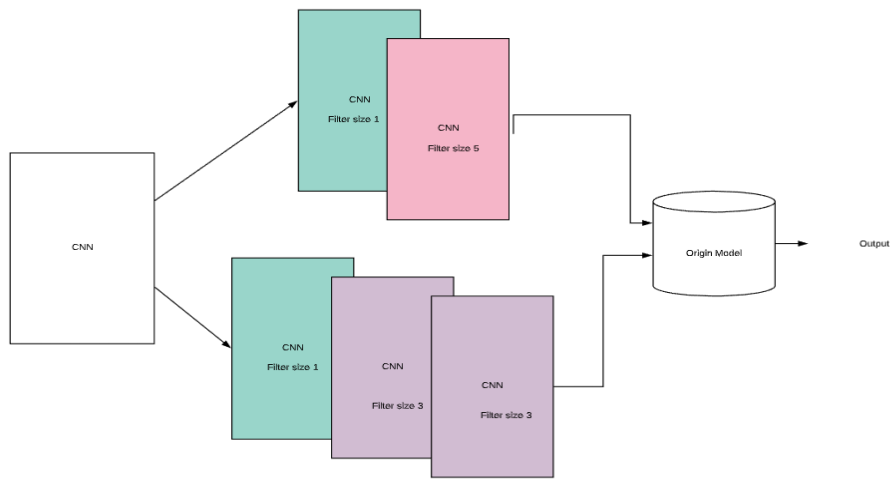

*Figure 3, CNN 2.0 architecture combined filter size 3 and 5*

We now use both filter size 3 and filter size 5 layers and combine them to pipe into the original model. The purpose is to use both layers with filter size 3 and 5 to check if the mixed filter size can help to improve accuracy.

*Table 7: Accuracy of classification after the step 3*

| Model | Accuracy | Recall | Precision | F1 |
|---|---|---|---|---|
| CNN 2.0 | 64.25 | 68.37 | 64.10 | 66.17 |

---

[3] We use inspiration from Lecture 11: Advanced CNN - Google Slides by S. Kim.

*Step 4,*

We explore the impact of batch size on the accuracy by comparing with the current size 64.

| Model | Accuracy | Recall | Precision | F1 |
|---|---|---|---|---|
| Batch size 16 | 61.02 | 64.04 | 61.02 | 62.49 |
| Batch size 32 | 63.90 | 67.07 | 62.77 | 64.85 |
| Batch size 64 (Current) | **64.41** | **68.28** | **64.26** | **66.21** |
| Batch size 128 | 64.26 | 68.34 | 64.13 | 66.17 |
| Batch size 256 | 64.15 | 68.10 | 64.02 | 65.10 |

*Table 8, Adjusted CNN 2.0 accuracy, step 4*

As we can see the batch size does not have a big impact on the accuracy. Our hypothesis that a larger batch size could let the model learn more examples at a time, thus tuning the weight based on more factors, works when the batch size is small. However, this approach does not work when the batch size > 64, since the accuracy is decreased.

*Step 5*

| Model | Accuracy | Recall | Precision | F1 |
|---|---|---|---|---|
| SGD | 37.80 | 37.04 | 36.74 | 36.89 |
| Adam | 64.41 | 68.28 | 64.26 | 66.21 |
| AdaBelief | 64.12 | 67.68 | 64.00 | 65.79 |

*Table 9, Adjusted CNN accuracy, attempt4*

Our current network is using Adam as our optimizer. On this step, we compared its performance with AdaBelief (Zhuang et.al., 2020) and Stochastic Gradient Descent. Both optimizers performer worse than Adam.

From the reported results, we observe that Recall is consistently higher than Precision. With its better Recall results, CNN 2.0 correctly identifies approx. 68% - 68.37 % patients of the appropriate class.

# 4.0 The Ensemble Model
## 4.1 Analysis on predicted values

As we have seen from the previous results, our accuracy is bottlenecked around 64%. To overcome the bottleneck, we analyze the class evaluation on each class by using the best model we have build. This is CNN 2.0 built on oversampling method ADASYN (Table 4, Sec. 2.4).

| **Measure** | **Readmission days** | **Results** |
|---|---|---|
| **Recall class 0** | 0 days | 55.01 |
| **Recall class 1** | < 30 days | 98.46 |
| **Recall class 2** | > 30 days | 51.38 |
| **Precision class 0** | 0 days | 62.98 |
| **Precision class 1** | < 30 days | 69.40 |
| **Precision class 2** | > 30 days | 60.40 |

*Table 10, class evaluation (%) on the model CNN 2.0-ADASYN*

We take the class evaluations from our best model with 64.41% Accuracy. As we can observe from the table, the Recall class 1 is around 98.45% which means we correctly labeled 98% of the instances in class 1. We can also see that the Precision of class 1 is also higher than the precision of class 2 and 0. This situation happens to all 10-fold cross-validation confusion matrices, as well as the confusion matrix on Gradient Boosting and Random Forest (Hu and Sokolova, 2020). In another word, our model performs well in classifying class 1.

To identify issues with the 0 and 2 class classification, we first assess the feature impact on the 0 and 2 classes classification. We have used Chi-square, Correlation, and ANOVA to analyze the feature importance for the classification. There are 10 common features that the three feature selection methods pick among the top 15 important features. We are going to use them in a further analysis.   Table 11 lists their means and standard variations in respect to the three classes.

| Feature Name | Class 0   (mean/var) | Class 1 (mean/var) | Class 2 (mean/var) |
| --- | --- | --- | --- |
| Diabetes | 0.072527 / 0.067269 | 0.094818 / 0.068905 | 0.090219 / 0.082082 |
| Emergency Room | 0.539381  / 0.248456 | 0.604200  / 0.199312 | 0.611995 / 0.237464 |
| Neoplasms | 0.041386 / 0.039675 | 0.028916 / 0.021694 | 0.022511 / 0.022005 |
| No_Med | 0.255823  / 0.190383 | 0.210309 / 0.137164 |  0.204681 / 0.162792 |
| ***Other type (Discharge)*** | ***0.099628 / 0.089705*** | ***0.059234 / 0.043509*** | ***0.047216 / 0.044988*** |
| Transfer (Admission) | 0.300932 / 0.210378 | 0.444508  / 0.203910 | 0.348553 / 0.227070 |
| Transfer.1 (Discharge) | 0.393027 / 0.238564 | 0.331725 / 0.183458 | 0.316200 / 0.216224 |
| Insulin | -0.980028  / 1.193263 | -0.862781 / 1.013230 | -0.914601 / 1.224610 |
| ***number_inpatient*** | ***0.392046  / 0.773391*** | ***1.091505 / 2.718667*** | ***0.846258 / 1.956131*** |
| time_in_hospital | 4.280210  / 8.847392 | 4.687134 / 8.195800 |  4.511645 / 8.963910 |

*Table 11, Feature analysis by different classes*

There are two features – other type (Discharge) and number_impatient - that have a large discrepancy on the mean values in respect to the classes. Other features exhibit similar mean values on the three classes. Thus, the features in isolation do not provide enough information for the classification assessment.

Our next step is to investigate how well the algorithm differentiates between classes 0 and 2. In the data, class 0 has 52337 instances, i.e. approx. 53% of the dataset, class 1 has 34649 instances, i.e.35% of the dataset, class 2 has 11066 instances, i.e., 11% of the dataset.  For the binary classification on class 0 and class 2, we randomly pick 11066 instances of class 2 first, then using NearMiss to under-sample class 0 with respect to class 2. After under-sampling, we have 11066 instances of class 2 and class 0 (the data set A). Then we used Gradient Boosting to perform binary classify class 0 and class 2 by using 10-fold cross-validation.[4]    The obtained accuracy is 79.31%.  Although the accuracy is high, we note that class 2 is randomly picked, class 0 is under-sampled using NearMiss with respect to the class 2.

This high accuracy will not retain if we randomly pick both class 0 and class 2, when Accuracy drops to 55%. Again, the high accuracy will not retain if we use a balanced dataset without under sampling (the data set B) when Accuracy drops to 68%. In this case, no under-sampling methods is used, thus the accuracy we have is an ordinary prediction by the model.   This shows that under-sampling technique helped us to achieve the high accuracy in this binary classification. Noted that we did not pick all the instances of class 2 since we think that the accuracy is enough for us to progress. Further analysis can be done using different sampling techniques.

---

[4] Note that Gradient Boosting outperformed other shallow Machine Learning algorithms on the same classification task (Hu and Sokolova, 2020).

### 4.2 The Gradient Boosting and CNN ensemble

We make use of the performance improvement we achieved above. We combine Gradient Boosting and CNN 2.0 and employ the results of Sec 4.1.

We use an oversampled (ADASYN) dataset, then train CNN on it, accept classification of class 1, and leave classification on classes 0 and 2 to Gradient Boosting. The dataset is generated as the data set B in the previous section and then we perform oversampling technique ADASYN on the dataset. The total dataset: 106880 instances, divided among three classes.

| Predicted/actual | 0 | 1 | 2 |
|---|---|---|---|
| 0 | 21112 | 97 | 13422 |
| 1 | 4316 | 25888 | 7417 |
| 2 | 12918 | 263 | 21447 |

*Table 12 The confusion matrix across 10-fold validation by CNN*

The confusion matrix above shows the confusion matrix across 10-fold cross-validation. In total, we have 97+25888+263 = 26248 prediction on class 1. We accept all the classification results for class 1. We then remove 26248 instances of class 1 from the data set. Total dataset: 80632 instances divided among two classes. Those 80632 instances have been classified by Gradient Boosting. The results are shown in Table 13.

| Predicted/actual | 2 | 0 |
|---|---|---|
| 2 | 21235 | 13130 |
| 0 | 12255 | 22279 |

*Table 13 Confusion matrix on binary classification by Gradient Boosting*

In total, from the original 106880 instances, CNN has correctly classified 25888 instances on class 1, followed by Gradient Boosting that correctly classified 43514 instances on classes 0 and 2. Thus, our accuracy is 64.94%. Compare to the best result we have by using CNN-ADASYN (Table 4, 64.41%), we improved Accuracy on > 0.5%.

*Table 14: Results of the CNN-Gradient Boosting ensemble.*

| Model | Accuracy | Recall | Precision | F1 |
|---|---|---|---|---|
| Ensemble | 64.94 | 68.87 | 65.04 | 61.56 |

## 5. Previous Work

Several works used Deep Learning algorithms to classify the patient re-admission stays. We listed results obtained on the same data set. Note that all the previous work has been done for binary classification.

Bhuvan et.al (2016), performed an additional cost analysis on patient's readmission and the binary prediction on if the patient will be readmitted in 30 days or would not. They concluded that Random Forest and Neural Networks are the best models with an area of 0.65 and 0.654 by using under precision-recall curve.

Hammoudeh et.al. (2018) binary classified would the patient be readmitted in 30 days or would not. The researchers used CNN and achieved an accuracy of 92%. They have applied an additional Z-norm to the binary encoded features, removed around 70,000 duplicate instances, and applied early stopping technique.

Li et.al (2018) achieved an accuracy of 88% for the binary classification of the readmission days. The team employed CFDL (Collaborative Filtering- enhanced Deep Learning) algorithm.

Avram et.al. (2020) correctly identified 82 % of the patients with diabetes. The authors used Deep Neural Networks (DNN).

Different from the listed work, we proposed a novel ensemble model that consists of a Convolutional Neural Networks and Gradient Boosting. Whereas the previous research has focused on the binary classification on readmission in 30 days, we worked on three-class readmission classification: 0, < 30, and > 30 days.

# 6. Conclusions and Future Work

In this work, we have reported application of Convolutional Neural Networks to multi-class classification of a large medical data set. Specifically, we investigated the impact of batch size, learning rate, and epoch by parameter tuning, and analyzed why CNN reached a bottleneck approx. 64.41%. We have proposed an ensemble model that consists of CNN and Gradient Boosting. The ensemble overcomes the accuracy bottleneck and improves classification performance with Accuracy of 64.93%, using 10-fold cross-validation.

In addition, we presented the results of different oversampling techniques. In our task, ADASYN outperforms the other methods. By comparing the results obtained by three optimizers, we have shown that Adam outperformed the other optimizers.

For future work on the ensemble model, we could apply NearMiss on each fold of the 10-fold cross-validation, perform binary classification with "0 days" and ">30 days", and then apply CNN on the entire fold for "<30 days". We could also migrate the binary classification to CNN by using the binary cross-entropy function. Another idea is that we could turn the multi-classification problem into three binary classification tasks, then analyze the results and label the test data with the most voted value. Also, RNN deserves more attention, as its parameter tuning has minor influence on accuracy around ~2%.

# References

Avram, R., Olgin, J., Kuhar, P., Hughes, J., Marcus, G.M., Pletcher, M., Aschbacher, K., & Tison, G. (2020). A digital biomarker of diabetes from smartphone-based vascular signals. Nature Medicine, 1-7.

Bhuvan, M.S., Kumar, A., Zafar, A., & Kishore, V. (2016). Identifying Diabetic Patients with High Risk of Readmission. ArXiv, abs/1602.04257.

Boehmke, B. C., Greenwell, B. (2020). Hands-on machine learning with R. Boca Raton ; London ; New York: CRC Press.

Hammoudeh, A. et al. "Predicting Hospital Readmission among Diabetics using Deep Learning." EUSPN/ICTH (2018).

Holzinger, A., Biemann, C., Pattichis, C. S., & Kell, D. B. (2017). What do we need to build explainable AI systems for the medical domain?. *arXiv preprint arXiv:1712.09923*.

Hosseini, M.M., Zargoush, M., Alemi, F. et al. Leveraging machine learning and big data for optimizing medication prescriptions in complex diseases: a case study in diabetes management. J Big Data 7, 26 (2020). https://doi.org/10.1186/s40537-020-00302-z

Hu, Y., and Sokolova, M. (2020) Explainable Multi-class Classification of Medical Data, deposited at arxiv.

Kingma, D. P. & Ba, J. (2014). Adam: A Method for Stochastic Optimization (cite arxiv:1412.6980Comment: Published as a conference paper at the 3rd International Conference for Learning Representations, San Diego, 2015)

LeCun, Y., Bengio, Y. & Hinton, G. Deep learning. Nature 521, 436–444 (2015). https://doi.org/10.1038/nature14539

Li, X., & Li, J. (2018). Health Risk Prediction Using Big Medical Data - a Collaborative Filtering-Enhanced Deep Learning Approach. 2018 IEEE 20th International Conference on e-Health Networking, Applications and Services (Healthcom), 1-7.

O'Shea, K., & Nash, R. (2015). An introduction to convolutional neural networks. *arXiv preprint arXiv:1511.08458.*

Reid, C. (2019). Diabetes Diagnosis and Readmission Risks Predictive Modelling: USA.

Shulan M, Gao K, Moore CD. Predicting 30-day all-cause hospital readmissions. Health Care Manag Sci. 2013 Jun;16(2):167-75. doi: 10.1007/s10729-013-9220-8. Epub 2013 Jan 27. PMID: 23355120.

Sokolova, M., & Lapalme, G. 2009. A systematic analysis of performance measures for classification tasks. Inf. Process. Manage. 45, 4, p.p. 427–437. DOI:https://doi.org/10.1016/j.ipm.2009.03.002

Strack, B., DeShazo, J., Gennings, C., Olmo, J., Ventura, S., Cios, K., Clore, J. "Impact of HbA1c Measurement on Hospital Readmission Rates: Analysis of 70,000 Clinical Database Patient Records", BioMed Research International, Article ID 781670, 2014. https://doi.org/10.1155/2014/781670

Verma, A. Pytorch [Basics]-Intro to RNN. 30 June 2020, towardsdatascience.com/pytorch-basics-how-to-train-your-neural-net-intro-to-rnn-cb6ebc594677.

Zhuang, J, et al. "AdaBelief Optimizer: Adapting Stepsizes by the Belief in Observed Gradients." ArXiv.org, 24 Oct. 2020, arxiv.org/abs/2010.07468.

# Appendix 1

*23 medication features, with values "Up", "Steady", "Down" and "No"*

Metformin, repaglinide, nateglinide, chlorpropamide, glimepiride, acetohexamide, glipizide, glyburide, tolbutamide, pioglitazone, rosiglitazone, acarbose, miglitol, troglitazone, tolazamide, examide, citoglipton, glyburide-metformin, glipizide-metformin, glimepiride-pioglitazone, metformin-rosiglitazone, metformin-pioglitazone

# Appendix 2

We applied Recurrent Neural Networks (RNN) to classify the data set and see if we can get better results. We first construct the simple RNN shown in the figure below.

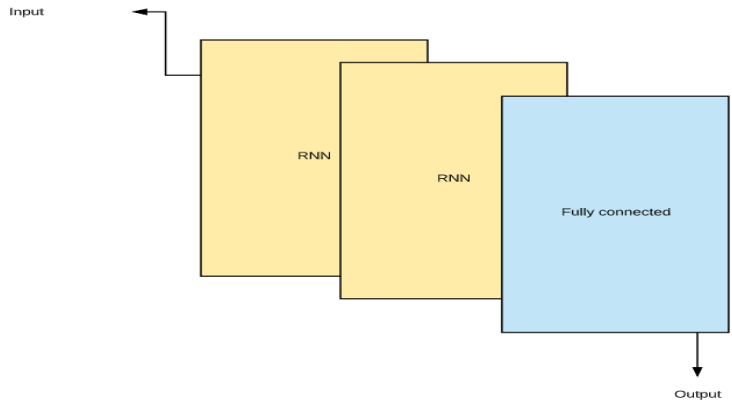

, Simple RNN architecture

We will test our RNN by different epochs and test if the accuracy is sensitive to the number of epochs. We are expecting the higher epochs leads to higher accuracy since more epochs will stabilize the weights in the networks.

| Model | Accuracy | Recall | Precision | F1 |
| --- | --- | --- | --- | --- |
| Epoch 50 (Current ) | 57.35 | 57.18 | 57.11 | 57.14 |
| Epoch 100 | 60.50 | 62.28 | 60.37 | 61.31 |
| Epoch 200 | 61.75 | 62.14 | 61.47 | 61.80 |
| Epoch 500 | 61.73 | 61.74 | 61.36 | 61.55 |

*Adjusted RNN accuracy*

We made a change to the number of epochs after we observed that when training on epoch 50, the loss value tends to decrease at the end of epochs. Thus we decided to train on more epochs. As we can see the accuracy did increase, but it still not better than 64.4%

*Performance with four layers of RNN*

| Model | Accuracy | Recall | Precision | F1 |
| --- | --- | --- | --- | --- |
| Epoch50 | 55.36 | 54.83 | 55.06 | 54.90 |